\documentclass{article}

\usepackage{docmute}

\usepackage{arxiv}

\usepackage[utf8]{inputenc}
\usepackage[T1]{fontenc}

\usepackage{amsmath}
\usepackage{amssymb}
\usepackage{amsfonts}
\usepackage{amsthm}
\usepackage{bm}
\usepackage{comment}
\usepackage{fancybox}
\usepackage{framed}
\usepackage{color}
\usepackage[dvipdfmx]{graphicx}
\usepackage{multicol}
\usepackage{multirow}
\usepackage{pdflscape}
\usepackage{verbatim}
\usepackage{subfigure}
\usepackage{cite}
\usepackage{hyperref}
\usepackage{url}

\hypersetup{
    colorlinks = true,
    urlcolor = blue,
    linkcolor = red,
    citecolor = green
}

\usepackage{algorithm}
\usepackage[noend]{algpseudocode}
\usepackage{algorithmicx}

\expandafter\def\expandafter\UrlBreaks\expandafter{\UrlBreaks
  \do\a\do\b\do\c\do\d\do\e\do\f\do\g\do\h\do\i\do\j%
  \do\k\do\l\do\m\do\n\do\o\do\p\do\q\do\r\do\s\do\t%
  \do\u\do\v\do\w\do\x\do\y\do\z\do\A\do\B\do\C\do\D%
  \do\E\do\F\do\G\do\H\do\I\do\J\do\K\do\L\do\M\do\N%
  \do\O\do\P\do\Q\do\R\do\S\do\T\do\U\do\V\do\W\do\X%
  \do\Y\do\Z}

\algrenewcommand\algorithmicindent{1.0em}

\algnewcommand\algorithmicforeach{\textbf{for each}}
\algdef{S}[FOR]{ForEach}[1]{\algorithmicforeach\ #1\ \algorithmicdo}

\algnewcommand\AlgAnd{\textbf{and} }
\algnewcommand\AlgOr{\textbf{or} }

\algrenewcommand\textproc{}

\algnewcommand{\Initialize}[1]{
	\State \textbf{Initialize:}
 	\State \hspace*{\algorithmicindent}\parbox[t]{0.8\linewidth}{\raggedright #1}}

\title{A Low-Cost Neural ODE with Depthwise Separable Convolution for Edge Domain Adaptation on FPGAs}

\author{
  Hiroki Kawakami\\
  Keio University\\
  3-14-1 Hiyoshi, Kohoku-ku, Yokohama, Japan\\
  \texttt{kawakami@arc.ics.keio.ac.jp}\\
  \And
  Hirohisa Watanabe\\
  Keio University\\
  3-14-1 Hiyoshi, Kohoku-ku, Yokohama, Japan\\
  \texttt{watanabe@arc.ics.keio.ac.jp}\\
  \And
  Keisuke Sugiura\\
  Keio University\\
  3-14-1 Hiyoshi, Kohoku-ku, Yokohama, Japan\\
  \texttt{sugiura@arc.ics.keio.ac.jp}\\
  \And
  Hiroki Matsutani \\
  Keio University\\
  3-14-1 Hiyoshi, Kohoku-ku, Yokohama, Japan\\
  \texttt{matutani@arc.ics.keio.ac.jp} \\
}

\begin{document}

\maketitle

\begin{abstract}
    High-performance deep neural network (DNN)-based systems are in high demand in edge environments.
    Due to its high computational complexity, it is challenging to deploy DNNs on edge devices with strict limitations on computational resources.
    In this paper, we derive a compact while highly-accurate DNN model, termed dsODENet, by combining recently-proposed parameter reduction techniques: Neural ODE (Ordinary Differential Equation) and DSC (Depthwise Separable Convolution).
    Neural ODE exploits a similarity between ResNet and ODE, and shares most of weight parameters among multiple layers, which greatly reduces the memory consumption.
    We apply dsODENet to a domain adaptation as a practical use case with image classification datasets.
    We also propose a resource-efficient FPGA-based design for dsODENet, where all the parameters and feature maps except for pre- and post-processing layers can be mapped onto on-chip memories.
    It is implemented on Xilinx ZCU104 board and evaluated in terms of
    domain adaptation accuracy, inference speed, FPGA resource utilization,
    and speedup rate compared to a software counterpart.
    The results demonstrate that dsODENet achieves comparable or slightly better domain adaptation accuracy compared to our baseline Neural ODE implementation, while the total parameter size without pre-
    and post-processing layers is reduced by 54.2\% to 79.8\%.
    Our FPGA implementation accelerates the inference speed by 23.8 times.
    \end{abstract}




\section{Introduction} \label{sec:intro}
To improve the accuracy of CNNs (Convolutional Neural Networks) in image recognition tasks, a typical approach is to build deeper models by stacking more convolutional layers~\cite{resnet}.
Although such image recognition tasks are in high demand in edge environments, computation resources are strictly limited in edge devices, 
making it difficult to use high-performance CNN models.
To reduce the amount of parameters and mitigate this issue, light-weight neural network models have been developed \cite{xception,mobilenets,Howard2019}.
Their key idea is to employ DSC (Depthwise Separable Convolution) that decomposes a conventional convolutional layer into two smaller convolutional steps.

ResNet \cite{resnet} is one of conventional CNN models
that stacks a lot of layers for a higher accuracy.
To reduce the parameter of ResNet, by utilizing a similarity to ODE (Ordinary Differential Equation), Neural ODE \cite{ode} repeatedly uses weight parameters instead of having a lot of different parameters.
More specifically, ResNet consists of sets of layers or building blocks.
An input to a building block is added to an output of the block via a shortcut connection for the residual learning.
This stacking structure of ResNet is interpreted as an ODE solver, and one execution of a building block is interpreted as one step of the ODE solver.
By repeatedly executing the same building block $C$ times instead of implementing $C$ different building blocks, the parameter size of these $C$ blocks in ResNet is theoretically reduced to $\approx 1/C$.
Thus, Neural ODE becomes significantly small compared to that of ResNet, and can be implemented in resource-limited edge devices.

Recently its implementation on a low-end FPGA (Field-Programmable Gate Array) device has been reported in \cite{watanabe}.
However, its performance improvement is limited since only one or two building blocks are implemented on the programmable logic, and it does not employ any other parameter reduction techniques.
For example, FPGA-based neural network accelerators and their optimization techniques, such as binarization and quantization, are surveyed in \cite{guo2018}.
In \cite{FMSCNN}, to fully exploit small but high-throughput on-chip memories of FPGAs,
a Feature-Map-Split-CNN technique that splits a feature map into smaller patches to be stored in the on-chip memories is proposed.
FPGA-optimized multipliers for DNNs that minimize information loss from quantization are studied in \cite{faraone2020}. 
DSC is applied to an FPGA-based CNN accelerator in \cite{Bai2018}. 
Binary neural network (BNN) is another approach to reduce memory sizes and computation costs.
In \cite{FracBNN}, FracBNN that exploits fractional activations to improve the accuracy of BNNs is proposed.

In this paper, a combination of Neural ODE and DSC, called dsODENet, is proposed and implemented for FPGAs to fully utilize on-chip memory resources.
As a practical use case, dsODENet is applied to domain adaptation, which is useful in a common edge AI deployment scenario.
When a trained model at server side is deployed to edge devices, the distribution difference between training data and inference data acquired at edge devices often causes a performance degradation, which can be dealt with domain adaptation techniques.
There are several forms of edge training scenarios, as surveyed in \cite{edge_in}.
In this paper, we assume an edge training scenario, where the edge training is done at edge servers (e.g., home servers and MEC servers) located between edge devices and cloud servers.
The edge server has a general teacher model and datasets collected at the edge environment, leading to improved scalability and security.
The edge server executes the proposed domain adaptation method so that it produces the student models optimized for the edge environment.
The students models are then used in FPGA-based edge devices.

Please note that our approach is basically orthogonal to quantization techniques and can be combined with them.
dsODENet is implemented on Xilinx ZCU104 board and evaluated in terms of the domain adaptation accuracy using image classification datasets, training speed, FPGA resource utilization, and speedup rate compared to a software execution
\footnote{This paper is an extended version of a conference version \cite{Kawakami_PDP22} by fully revising the FPGA implementation. The demonstration video of the revised FPGA implementation is available at \url{https://youtu.be/54EFQ6ZuX9c}.}.

The rest of this paper is organized as follows.
Section \ref{sec:related} introduces baseline technologies behind our proposal.
Section \ref{sec:proposal} introduces our domain adaptation method and
Section \ref{sec:design} proposes dsODENet and describes the FPGA implementation of dsODENet.
Section \ref{sec:eval} shows evaluation results and
Section \ref{sec:conc} concludes this paper.


\section{Related Work} \label{sec:related}

\subsection{Depthwise Separable Convolution}\label{subsec:DSC}
\begin{figure}[ht]
    \centering
    \includegraphics[keepaspectratio, width=1\linewidth]{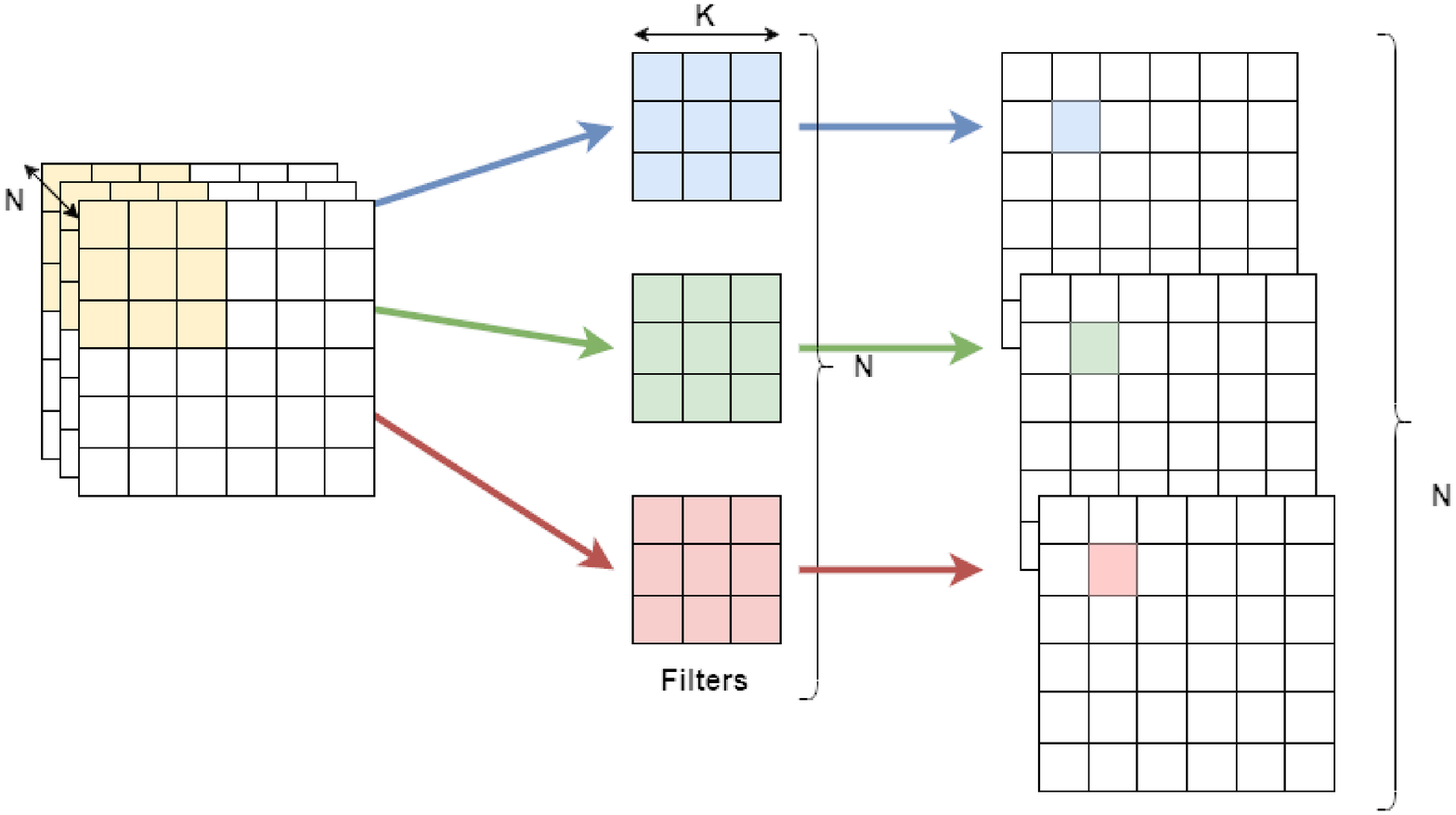}
    \caption{Depthwise Convolutional step}
    \label{fig:depth}
\end{figure}

\begin{figure}[ht]
    \centering
    \includegraphics[keepaspectratio, width=1\linewidth]{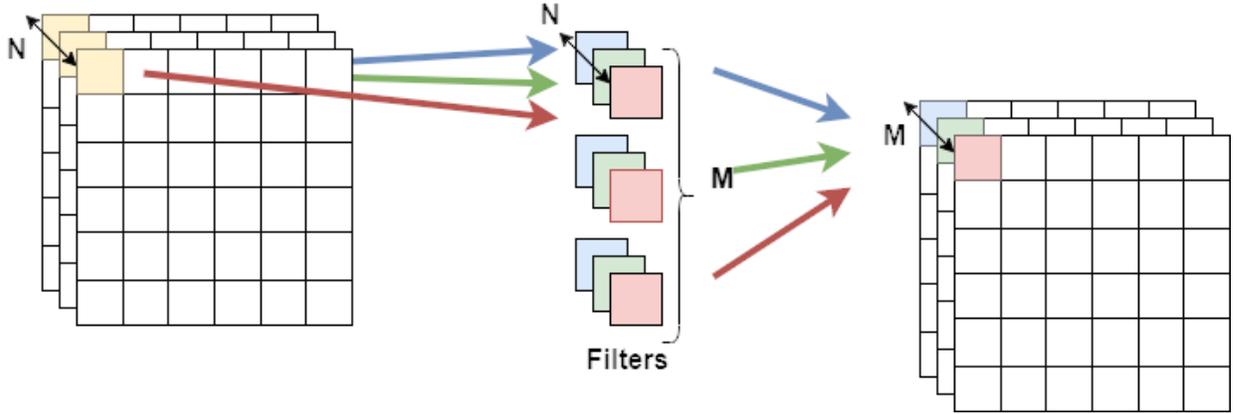}
    \caption{Pointwise Convolutional step}
    \label{fig:point}
\end{figure}

CNNs typically stack a set of convolutional layers for a higher image recognition accuracy, and each convolutional layer contains a lot of parameters.
The parameter size of a convolutional layer can be calculated as a product of the number of input channels, the number of output channels, and kernel size.
Let $N$, $M$, and $K$ be the number of input channels, the number of output channels, and the length of one side of kernel, respectively.
The weight parameter size of a conventional convolutional layer is $NMK^2$.
DSC decomposes a conventional convolutional layer into two smaller convolutional steps:
depthwise convolutional step and pointwise convolutional step (see Figures \ref{fig:depth} and \ref{fig:point}).

In DSC, as depthwise convolutional step, as shown in Figure \ref{fig:depth},
a convolution operation involving only spatial direction (the size is $K^2$) is applied for each of an input feature map.
Different weight parameters are used for each of $N$ input channels;
thus its weight parameter size is $NK^2$.
Then, an output feature map of the depthwise convolutional step 
(Figure \ref{fig:depth} right)
is fed to the pointwise convolutional step as an input.
As shown in Figure \ref{fig:point}, a $1\times 1$ convolution operation is applied for each of the input feature map and for each of $M$ output channels; thus its weight parameter size is $NM$.
The weight parameter size of DSC is $NK^2 + NM$ in total, which is approximately $K^2$ times reduction, assuming that $N, M \gg K$.


\subsection{Ordinary Differential Equation}\label{subsec:ode}
ODE is an ordinary differential equation that contains an unknown
function and its derivatives.
An example of a first-order differential equation is shown in Equation
\ref{eq:ode}.
\begin{align}\label{eq:ode}
    \frac{dz}{dt} = f(z(t), t, \theta ) ,
\end{align}
where $f(\cdot)$ is a known function and $\theta$ is the parameter.
When the initial value $z(t_0)$ is given, a problem to find $z(t_1)$ that
satisfies Equation \ref{eq:ode} is known as an initial value problem.
The solution is formulated as shown in Equation \ref{eq:ivp}.
\begin{align}\label{eq:ivp}
    z(t_1) &= z(t_0) + \int_{t_0}^{t_1}f(z(t), t, \theta)dt
\end{align}
In the right side of Equation \ref{eq:ivp}, the second term contains
an integral of the function, and thus it cannot be solved analytically
for arbitrary functions.
To solve the solution approximately, the following ODESolve function
is introduced.
\begin{align}\label{eq:odesolve}
    z(t_1)  &= ODESolve(\bm{z}(t_0), t_0, t_1, f)
\end{align}
As an ODESolve function, Euler method shown in Equation \ref{eq:euler}
can be used.
\begin{align}\label{eq:euler}
    z(t_{i+1}) &= z(t_i) + hf(z(t_i), t_i, \theta)
\end{align}
Euler method is a first-order approximation for solving the initial
value problem, and it is used in this paper.

\subsection{Neural ODE} \label{subsec:neural_ode}
ResNet is a well-known neural network architecture that can increase
the number of stacked layers or building blocks by introducing
shortcut connections.
Using a shortcut connection, an input feature map to a building block
is temporarily saved and then it is added to the original
output of the building block to generate the final output of the block.
Let $z_t$, $\theta$, and $f(z_t, \theta _t)$ be an input feature map
to a building block, parameters of the building block, and processing
result of the block, respectively.
The final output of the building block is represented as Equation
\ref{eq:resnet}.
\begin{align}\label{eq:resnet}
    \bm{z}_{t+1}=f(\bm{z}_t,\theta_t)+\bm{z}_t
\end{align}
Please note that Equation \ref{eq:resnet} is similar to Equation
\ref{eq:euler}, though the former basically assumes vector values
while the latter assumes scalar values.
Based on this similarity, one building block can be interpreted as one
step of an ODESolve function.
Assuming that Euler method is used as an ODE solver, it can be interpreted that a first-order approximation is applied to solve the output of the building block.
In this paper, one building block is called ODEBlock and the whole
network architecture consisting of ODEBlocks is called ODENet.

\begin{figure}[ht]
\centering
\includegraphics[keepaspectratio, width=1\linewidth]{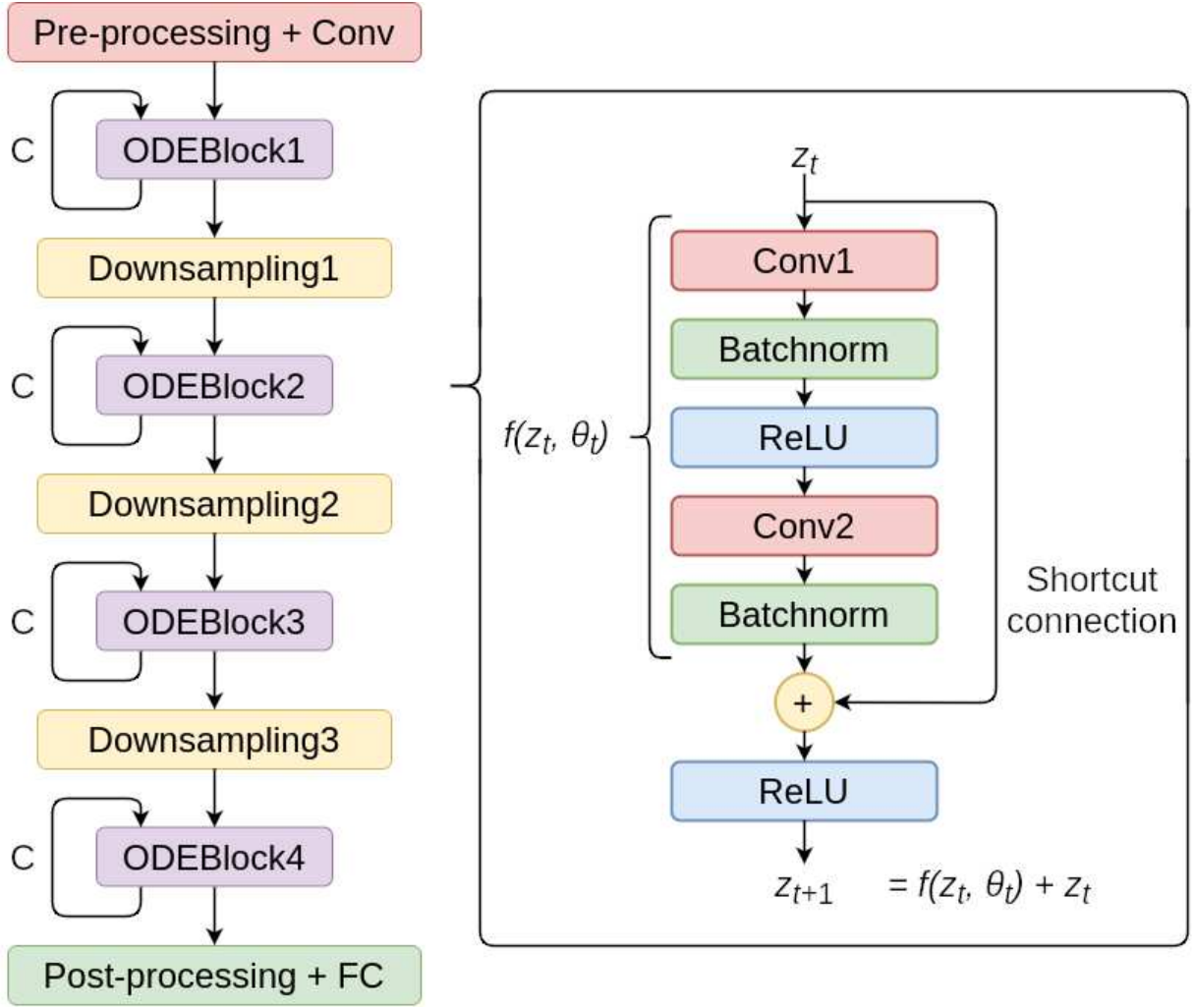}
\caption{ODENet architecture}
\label{fig:odenet}
\end{figure}

Figure \ref{fig:odenet} illustrates a practical example of ODENet.
The left side of Figure \ref{fig:odenet} shows the overall ODENet
architecture that consists of ODEBlocks and downsampling blocks in
addition to pre-processing and post-processing layers.
The pre-processing layer has a convolutional layer (denoted as Conv)
and the post-processing layer has a fully-connected layer (denoted as FC).
The right side of Figure \ref{fig:odenet} shows the internal structure
of an ODEBlock that consists of convolutional layers, batch
normalization layers (denoted as Batchnorm), and ReLU layer.
Each ODEBlock is repeatedly executed $C$ times in
ODENet, while in ResNet, $C$ different building blocks are executed
once.
Let $O(L)$ be the parameter size of one building block or ODEBlock.
Total parameter sizes of convolutional layers in ResNet and ODENet are $O(CL)$ and $O(L)$,
respectively; thus ODENet can significantly reduce the parameter size.

\subsection{Edge Domain Adaptation}\label{subsec:domain}
Domain adaptation is a kind of transfer learning, where knowledge
obtained at a source domain is transferred to a different domain called target domain.
It is typically assumed that the source domain has enough labeled training data while the target domain does not.
There are three domain adaptation approaches: discrepancy-based approach, adversarial-based approach, and reconstruction-based approach.
The discrepancy-based approach uses fine-tuning to adapt to a target domain by reducing the domain shift.
The adversarial-based approach reduces the distance between source and target distributions by using a discriminator to distinguish between source and target domains.
The reconstruction-based approach assumes that reconstruction can help improving the performance of domain adaptation.

As explained in Section \ref{sec:intro}, it is useful in a common edge AI deployment scenario.
MobileDA \cite{MobileDA} is a discrepancy-based domain adaptation technique for edge
devices based on knowledge distillation and DeepCORAL \cite{deep_coral}.
Knowledge distillation is a learning method that uses a large model and a small model.
The large model is called a teacher model and the small model is called a student model.
By transferring knowledge from a teacher model to a student model, the student model can acquire a similar accuracy to the teacher model.
As a result, the model can be made smaller while keeping the accuracy.
DeepCORAL is a method to reduce the distance between domains.
In MobileDA, knowledge distillation and DeepCORAL are used when training student models, which will be detailed in the next section.

Since the target domain is an edge environment in the edge domain
adaptation scenario, the target model should be further reduced in
parameter size and computation cost.
Although pruning, quantization, and distillation are very common model
compression techniques, in this paper we propose a combination of
ODENet (see Figure \ref{fig:odenet}) and DSC
to further reduce the number of parameters of the target domain model.




\section{Domain Adaptation Method} \label{sec:proposal}

In this paper, a modified version of MobileDA is used as an edge
domain adaptation procedure to gain a higher performance.
While MobileDA uses one teacher model and one student model as knowledge distillation, our approach uses one teacher model and two student models.
ResNet is used as the teacher model, while the combination of ODENet
and DSC, called dsODENet, is used in the two student models.
In Section \ref{sec:eval}, our approach will be compared to the original MobileDA.

\begin{figure*}[ht]
    \centering
    \includegraphics[keepaspectratio, width=1\linewidth]{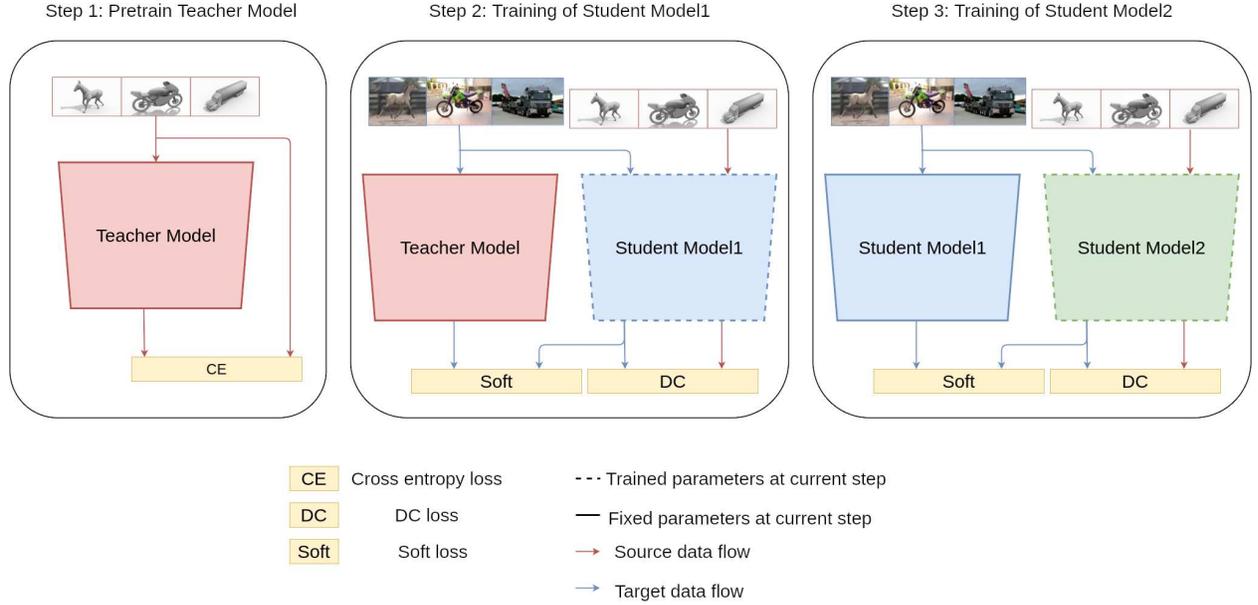}
    \caption{Training steps of our approach.\ \ 
    Step1: Training a teacher model on a server (left).
    Step2: Training a student model1 on an edge device based on the teacher model with fixed parameters (middle).
    Step3: Training a student model2 on an edge device based on the student model1 with fixed parameters (right).
    }
\label{fig:prop_method2}
\end{figure*}

The training phase consists of three steps.
First, a teacher model is trained with source domain data in Step 1,
and then two student models are trained in Steps 2 and 3.
Finally, the student model makes inferences.
Steps 1, 2, and 3 of the learning process are shown in Figure
\ref{fig:prop_method2}.
In Step 2, the teacher model is fixed (i.e., no update is allowed),
and parameters of student model1 are trained by the help of the
teacher model.
In Step 3, the student model1 is fixed, and parameters of student
model2 are trained by the help of the student model1.
Step 3 is optional, and either student model1 or model2 (whichever shows better accuracy) can be
used for the prediction.

As shown in Figure \ref{fig:prop_method2}, two loss functions are used
in this domain adaptation: $L_{Soft}$ and $L_{DC}$.
$L_{Soft}$ is a soft target loss of knowledge distillation and it is
defined in Equation \ref{eq:Soft_Loss}.
$L_{DC}$ is a loss function borrowed from DeepCORAL \cite{deep_coral}
and it is defined in Equation \ref{eq:DC_Loss}.
\begin{align}\label{eq:Soft_Loss}
    L_{Soft} &= \mathbb{E}_{x_t\sim X_t}\sum_k [L(M_T(x_t), M_S(x_t))]
\end{align}
\begin{align}\label{eq:DC_Loss}
    L_{DC} = \frac{1}{4d^2}||C_s-C_t||^2_F,
\end{align}
where $L(\cdot)$ is a loss function, $M_S(x_t)$ is an output when
target domain data is fed to a student model, $M_T(x_t)$ is an output
when target domain data is fed to a teacher model, $C_s$ is a
covariance matrix of $M_S(x_s)$, $C_t$ is a covariance matrix of
$M_S(x_t)$, $d$ is degree of the covariance matrix (e.g., the number
of samples), and $||\cdot||^2_F$ is Frobenius norm, respectively.
Given that target domain labels are produced by a teacher model,
$L_{Soft}$ is a loss value computed by comparing the generated target
domain labels and those predicted by a student model.
$L_{DC}$ is computed by the distance between the covariance matrices
of the two domains.
The final loss function combines $L_{Soft}$ and $L_{DC}$ as shown in
Equation \ref{eq:Loss}.
\begin{align}\label{eq:Loss}
    L =& L_{Soft}+\lambda L_{DC}
\end{align}
$L_{DC}$ is weighted by a hyper-parameter $\lambda$ that controls the
strength of domain confusion.
A smaller $\lambda$ increases the importance of class
prediction results by a teacher model, which was trained by the source
domain data.
On the other hand, a larger $\lambda$ increases the importance of
domain invariant representation, but class prediction accuracy may be
weaken.

Here, target domain samples to be trained are selected by a
given threshold value.
Specifically, target domain samples are first fed to the teacher
model in Step 2 or the student model1 in Step 3.
Softmax function is then applied to the class prediction results so
that the sum of the probability of each class is 1.0.
If the highest class probability value of a sample is greater than a
given threshold value, the sample is used for the student model training.
This can prevent situations that incorrect labels
produced by the teacher model in Step 2 or student model1 in Step 3 are used for the student model training.
The overall flow of our domain adaptation method is summarized in Algorithm \ref{alg:prop_method}. 
\begin{algorithm}[ht]
    \caption{Proposed domain adaptation method}
    \label{alg:prop_method}
    \begin{algorithmic}
        \State {\textbf{Pretrain}: Training of teacher model}
        \For{each epoch}
            \State {1) Obtain $x_t$ from teacher model if the highest predicted probability value is higher than threshold}
            \State {2) Calculate the Soft Target Loss (Equation \ref{eq:Soft_Loss})}
            \State {3) Calculate the DC Loss (Equation \ref{eq:DC_Loss})}
            \State {4) Train student model1 by the loss function (Equation \ref{eq:Loss})}
        \EndFor
        \For{each epoch}
            \State {1) Obtain $x_t$ from student model1 if the highest predicted probability value is higher than threshold}
            \State {2) Calculate the soft target loss (Equation \ref{eq:Soft_Loss})}
            \State {3) Calculate the DC loss (Equation \ref{eq:DC_Loss})}
            \State {4) Train student model2 by the loss function (Equation \ref{eq:Loss})}
        \EndFor
        \State {\textbf{Output}: Student model1 and student model2}
    \end{algorithmic}
\end{algorithm}


\section{dsODENet}\label{sec:design}
\subsection{Models}\label{subsec:model}
\begin{figure}[ht]
\centering
\includegraphics[keepaspectratio, width=1\linewidth]{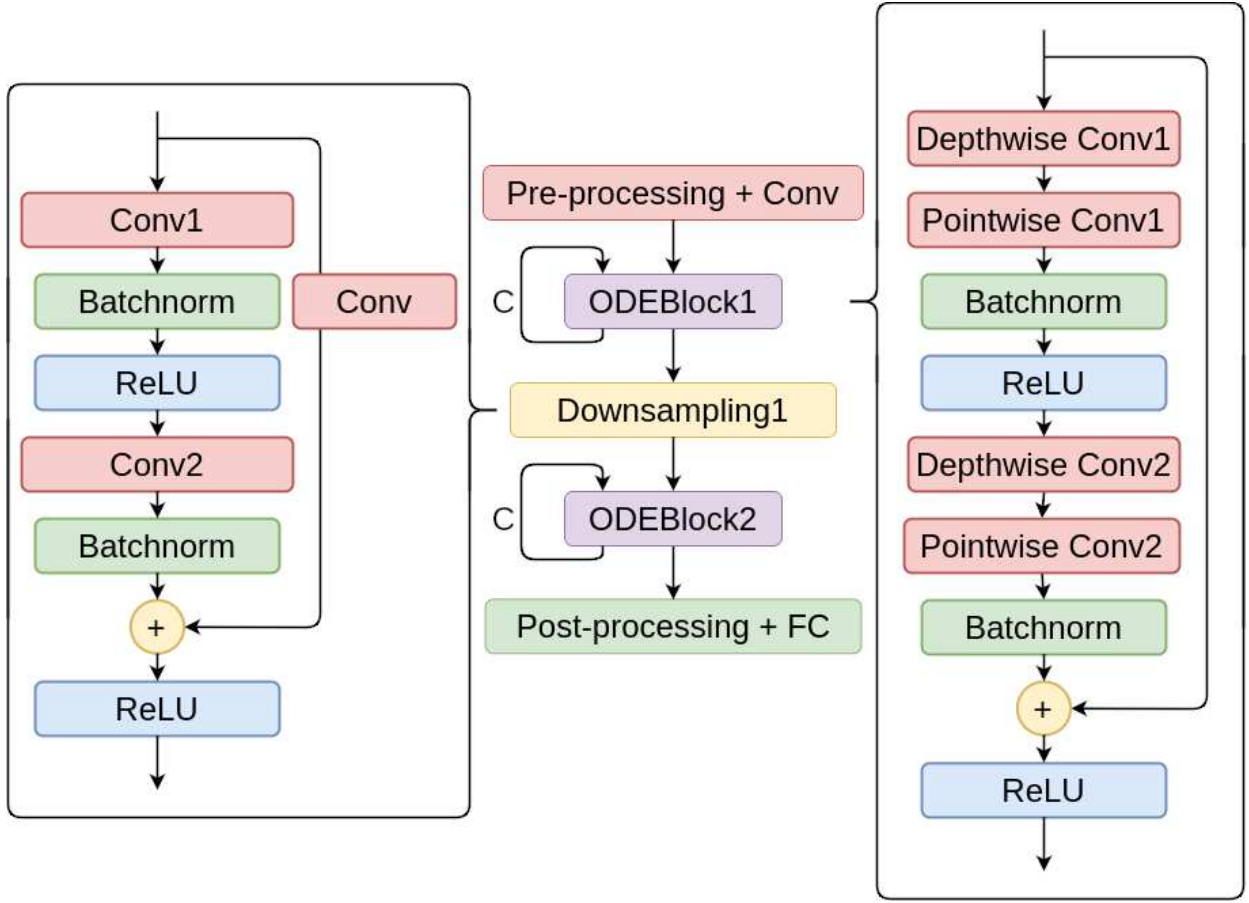}
\caption{Model with two ODEBlocks}
\label{fig:separable_ode2}
\end{figure}

\begin{figure}[ht]
\centering
\includegraphics[keepaspectratio, width=1\linewidth]{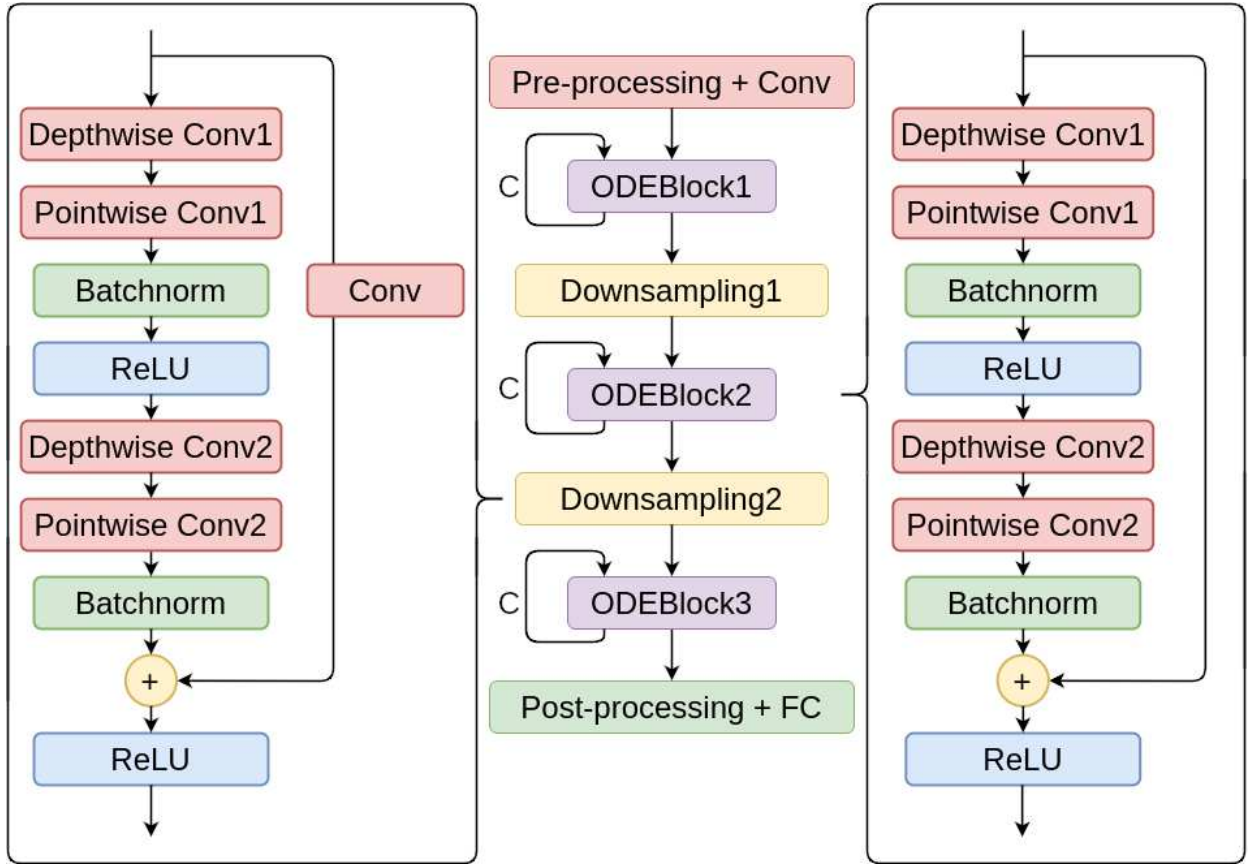}
\caption{Model with three ODEBlocks}
\label{fig:separable_ode3}
\end{figure}

Here, we propose a resource-efficient and lightweight DNN model, termed dsODENet,
that takes advantages of both ODENet and DSC for resource-limited FPGAs.
Figures \ref{fig:separable_ode2} and \ref{fig:separable_ode3} show two
dsODENet models: model with two ODEBlocks and that with three ODEBlocks.

In Figure \ref{fig:separable_ode2}, the right box shows an internal
structure of ODEBlocks and the left box shows that of a downsampling
block.
The structures of ODEBlocks and downsampling blocks are similar, but
in the ODEBlocks, input and output feature map sizes are the same and
$M=N$, while in downsampling block, input feature map size is scaled
down to $1/2 \times 1/2$ and $M=2N$.
Each ODEBlock is executed $C$ times, while the downsampling block is
executed once.
In the downsampling blocks, a $1\times 1$ convolutional operation with
stride 2 (denoted as Conv) is additionally applied to the shortcut connection.
To reduce the parameter size of the two ODEBlocks case, DSC can be
applied to convolutional layers of ODEBlocks and downsampling block.
We observed that the accuracy is sometimes sensitive to the DSC on the downsampling block that rescales the feature map.
Considering the stability, in the two ODEBlocks case, DSC is applied
to the two ODEBlocks while it is not applied to the downsampling
block, as shown in Figure \ref{fig:separable_ode2}.
Assuming that $N$ is 64 in ODEBlock1, total parameter size of ODEBlock1,
downsampling1, and ODEBlock2 in the ODENet without DSC 
is 598,016.
In dsODENet, the total size is 273,792, which is 54.2\% reduction.
Assuming a 32-bit fixed-point representation, their sizes are 19.1Mb
and 8.8Mb, respectively.


As shown in Figure \ref{fig:separable_ode3}, the three ODEBlocks case
has three ODEBlocks and two downsampling blocks.
DSC is applied to all the three ODEBlocks.
Regarding the downsampling blocks, their parameter sizes without DSC
are 221,184 and 884,736, respectively, assuming that $N$ is 64 in
ODEBlock1.
Their sizes depend on the numbers of input and output channels, and
these numbers are doubled once the downsampling is applied; thus, the
second downsampling block (Downsampling2) is much larger than that of
the first one (Downsampling1).
Since DSC on the downsampling block is sometimes sensitive to accuracy
as mentioned above, in this paper DSC is applied only to the second
downsampling block.
Please note that the total parameter size of ODEBlock1, Downsampling1,
ODEBlock2, Downsampling2, and ODEBlock3 in the ODENet without DSC 
is 2,695,168.
In dsODENet, 
the total parameter size is 544,000, which is 79.8\% reduction.
When a 32-bit fixed-point representation is employed, their sizes are 86.2Mb and 17.4Mb, respectively.
This parameter size reduction by DSC is significant since these
parameters can be implemented on BRAM or URAM of modest FPGA devices for
simplicity.

In Section \ref{sec:eval}, ResNet, ODENet, and dsODENet are used as student models of the domain adaptation as mentioned in Section \ref{sec:proposal}.
In this paper, ResNet-50 \cite{resnet} is used as a baseline.
$C$ is set to 10 so that the number of convolutional layers of ODENet/dsODENet is same as that of ResNet-50.

\subsection{FPGA Accelerator}\label{subsec:implementation}
%
This section describes the design of a dsODENet accelerator using FPGA SoC. 
As mentioned in Section \ref{sec:proposal}, dsODENet repeatedly executes ODEBlocks, which is a bottleneck in the execution time.
Thus, all the ODEBlocks and intermediate processing between them are accelerated by the programmable logic of the FPGA.

\subsubsection{Top Module}\label{subsubsection:overview_design}
\begin{figure}[ht]
    \centering
    \includegraphics[keepaspectratio, width=1\linewidth]{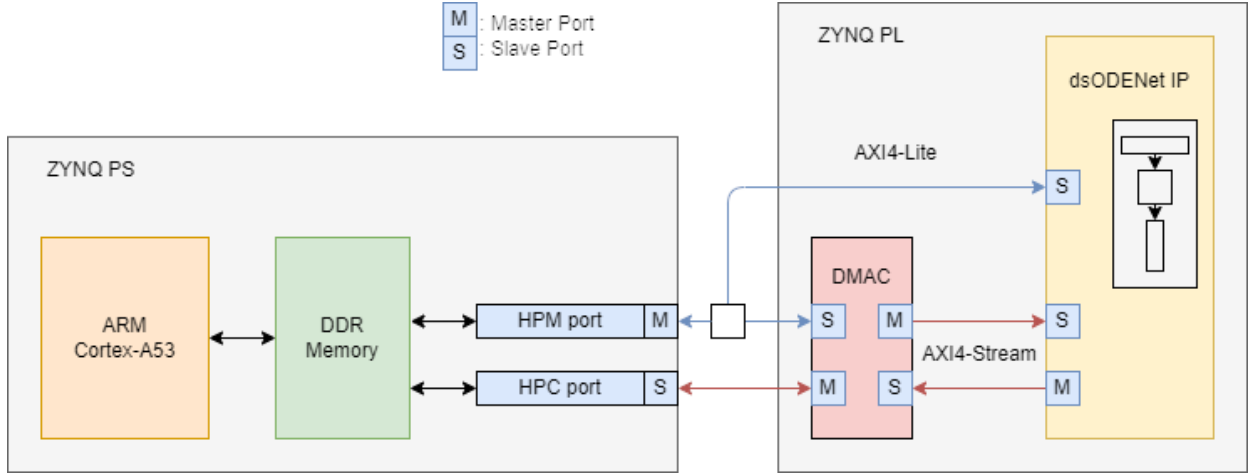}
    \caption{Board-level implementation}
    \label{fig:bord-level_implementation}
\end{figure}

%
%
Figure \ref{fig:bord-level_implementation} illustrates a block diagram of the board-level implementation, which is divided into a processing system (PS) part and a programmable logic (PL) part. 
The proposed dsODENet IP core and a direct memory access (DMA) controller are instantiated in the PL part. 
The PS part sets up the IP core and transfers input images to the PL part by the DMA controller, while the PL part computes feature maps of the incoming images sent from the PS part.
The PS part is also in charge of pre-processing and post-processing parts of dsODENet.
For high-speed data transfer, the DMA controller is connected to a 32-bit wide high-performance slave port (HPC port) in the AXI4-Stream protocol (red line in Figure \ref{fig:bord-level_implementation}). 
The control registers are connected to a high-performance master port (HPM port) via the AXI4-Lite interface (blue line in Figure \ref{fig:bord-level_implementation}).

The dsODENet core has two modes: weight transfer mode and feature map computation mode. 
In the weight transfer mode, the IP core receives weight parameters of dsODENet via the AXI4-Stream interface and stores them in on-chip BRAM and URAM buffers. 
The transfer mode is finished when the IP core sends back a 32-bit nonzero value as an acknowledge message.
In the feature map computation mode, a pre-processed input image ($\mathbb{R}^{8\times8\times3}$) is sent to the IP core, 
and then it is processed by ODEBlock1, Downsampling1, ODEBlock2, Downsampling2, and ODEBlock3 in sequence to produce the final feature map ($\mathbb{R}^{1\times1\times256}$).

These ODEBlocks are iteratively executed by feeding back the output of an iteration step to the input of the next step.
Since the weight parameters used in the iterations are the same, utilizations of internal memories (BRAM and URAM) can be significantly reduced.
In addition, the design is scalable in terms of the number of iterations, since the hardware resources are constant even if the number of iterations is increased.
Please note that the weight parameters and feature maps are packed in the internal memories by quantizing them with 24-bit or 20-bit fixed-point formats, as analyzed in Section \ref{subsubsec:quantization}.
By the quantization, these blocks are running efficiently without accessing on-board DRAM modules.

\subsubsection{dsODENet and ODEBlock Modules}\label{subsubsec:module}

\begin{figure}[ht]
    \centering
    \includegraphics[keepaspectratio, width=0.7\linewidth]{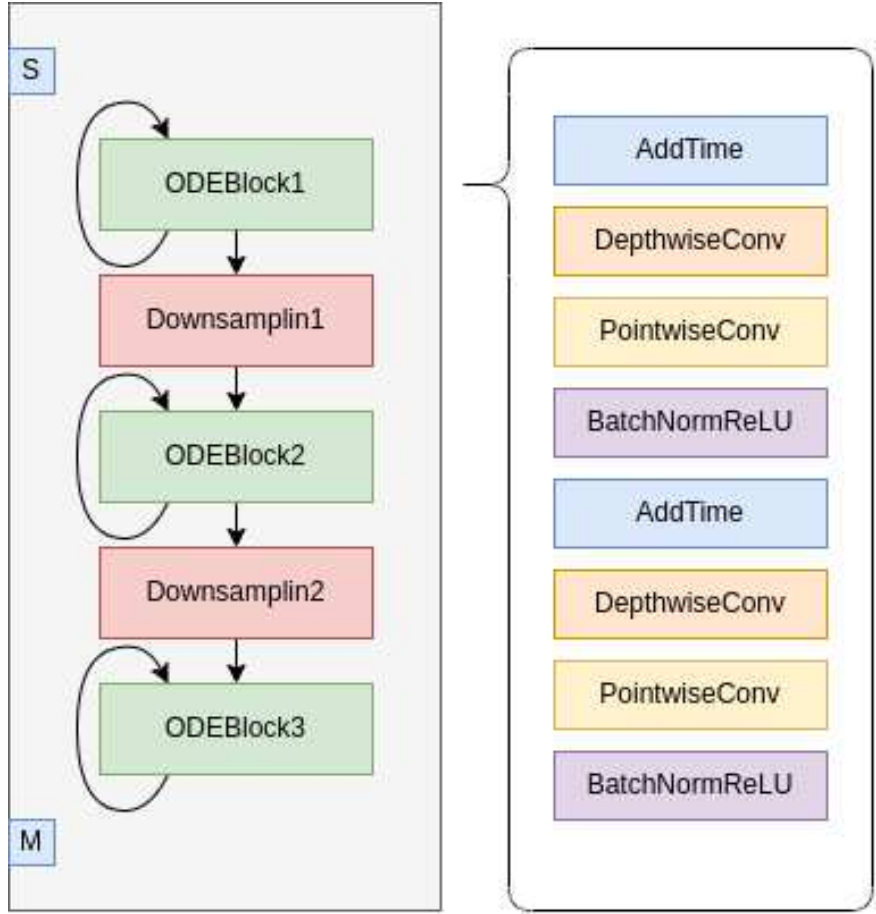}
    \caption{computation of dsODENet IP core}
    \label{fig:dsODENet_IPcore}
\end{figure}

As shown in Figure \ref{fig:dsODENet_IPcore}, the dsODENet IP core consists of five major modules: 
ODEBlock1, Downsampling1, ODEBlock2, Downsampling2, and ODEBlock3. 
The number of output channels is increased while the output feature map size (width and height) is decreased at each Downsampling block.
That is, output feature map sizes of ODEBlock1, Downsampling1, ODEBlock2, Downsampling2, and ODEBlock3 are $8 \times 8 \times 64$, $4 \times 4 \times 128$, $4 \times 4 \times 128$, $2 \times 2 \times 256$, $2 \times 2 \times 256$, respectively. 
As shown in Figure \ref{fig:dsODENet_IPcore}, each ODEBlock consists of two sets of AddTime, DepthwiseConv, PointwiseConv, BatchNorm, and ReLU modules.
AddTime module inserts a new channel representing the current iteration count to the feature map.
That is, it increases the number of channels by one, and its feature map buffer is also extended to store one more channel.
Since ODEBlocks have a skip-connection structure, three feature map buffers are implemented in each ODEBlock: two for input and output buffers of the convolutional modules and one for saving the original input feature map.

%
Loop unrolling and loop pipelining directives are used in DepthwiseConv, PointwiseConv, ReLU, and BatchNorm modules.
The unrolling factor is set to eight in these modules, so they are processed by eight arithmetic units in parallel. 

%
%
\subsubsection{Implementation}\label{subsubsec:implementation_details}
%
dsODENet IP core was designed and implemented with Xilinx Vitis HLS 2020.2 and Xilinx Vivado 2020.2 for the high-level synthesis and place-and-route of the design. 
We chose Xilinx Zynq UltraScale+ MPSoC series as a target FPGA platform, and ZCU104 evaluation board kit (XCZU7EV-2FFVC1156) was used in this paper.
The CPU on the FPGA device is ARM Cortex-A53 running at 1.2GHz.
We used Python 3.8.2 and PyTorch 1.10.2 in the software stack.

%
The weight parameters of dsODENet are trained by a domain adaptation method described in Section \ref{sec:proposal}.
The trained parameters are then transferred to the DRAM of the target FPGA SoC.
To start an inference processing at the PL part, the parameters are moved to internal BRAM and URAM buffers in the weight transfer mode.
In the feature map computation mode, 
an input feature map is transferred to the PL part, it is processed at the PL part, and then the output feature map is sent back to the PS part.
Please note that the parameters and feature maps are implemented on the BRAM and URAM of the FPGA to fully enjoy benefits of using fast on-chip memories.
A larger batch size is typically used in the training phase for a fast training.
In contrast, this inference accelerator assumes a single batch size since the inference processing is triggered by a per-request basis.


%
Although the BRAM and URAM sizes are 11Mb and 27Mb in total, respectively, their instance sizes are 36kb and 288kb, 
which means that, depending on the number and sizes of parameter arrays, a part of BRAM and URAM instances is underutilized. 
In our design, each parameter array is carefully implemented on either BRAM or URAM instances to minimize the underutilized on-chip memories.
%
More specifically, in the three ODEBlocks case, parameter arrays of 
normal convolutional layers of Downsampling1, those of depthwise and pointwise convolutional layers of Downsampling2, and those of pointwise convolutional layers of ODEBlock3 are implemented on the URAM instances; 
and the others are implemented on BRAM instances.




\section{Evaluations}\label{sec:eval}
The proposed dsODENet for FPGAs is evaluated with an edge domain
adaptation scenario using image recognition datasets.
For accuracy evaluations, it is implemented with Pytorch 1.8.1 and
torchvision 0.9.1.
For resource utilization and performance evaluations, it is
implemented with Xilinx Vivado v2020.2 for ZCU104 FPGA board.
Specification of the board is shown in Table \ref{tbl:zcu104}.
\begin{table}[ht]
    \centering
    \caption{Specification of ZCU104 board}
    \label{tbl:zcu104}
    \begin{tabular}{l|c} \hline
        OS   & Pynq Linux (Ubuntu 18.04)\\
        CPU  & ARM Cortex-A53 @ 1.2GHz\\ 
        DRAM & DDR4 2GB\\
        FPGA & Zynq UltraScale+ XCZU7EV-2FFVC1156\\ \hline
    \end{tabular}
\end{table}

\subsection{Number of Parameters}\label{subsec:parameters}
A comparison of the number of parameters for ResNet, ODENet, and dsODENet is shown in Tables \ref{tbl:parameters1} and \ref{tbl:parameters2}.
The percentage compared to the number of parameters in ResNet is listed in ODENet and dsODENet.
The reduction in the number of parameters from the third building block, which has a large number of parameters in ResNet, to ODEBlock3 in dsODENet is significant.
In total, dsODENet reduces the number of parameters of these blocks by 96.7\% compared to ResNet.
\begin{table*}[ht]
    \begin{minipage}[t]{.4\textwidth}
        \centering
        \caption{Number of parameters of ResNet}
        \label{tbl:parameters1}
        \begin{tabular}{l|l|r} \hline
                                             &        & \multicolumn{1}{c}{ResNet}\\ \hline
            \multirow{2}{*}{Building block1} & Conv1  & 368,640    \\
                                             & Conv2  & 368,640    \\ \hline
            \multirow{3}{*}{Downsampling1}   & Conv   & 8,192      \\
                                             & Conv1  & 73,728     \\
                                             & Conv2  & 147,456    \\ \hline
            \multirow{2}{*}{Building block2} & Conv1  & 1,474,560  \\
                                             & Conv2  & 1,474,560  \\ \hline
            \multirow{3}{*}{Downsampling2}   & Conv   & 32,768     \\
                                             & Conv1  & 294,912    \\
                                             & Conv2  & 589,824    \\ \hline
            \multirow{2}{*}{Building block3} & Conv1  & 5,898,240  \\
                                             & Conv2  & 5,898,240  \\ \hline
            Others                           &        & 9,728      \\ \hline
            Total                            &        & 16,639,488 \\ \hline
        \end{tabular}
    \end{minipage}
    \begin{minipage}[t]{.55\textwidth}
        \centering
        \caption{Numbers of parameters of ODENet and dsODENet}
        \label{tbl:parameters2}
        \begin{tabular}{l|l|rr|rr} \hline
                                           &        & \multicolumn{2}{c|}{ODENet} & \multicolumn{2}{c}{dsODENet}\\ \hline
            \multirow{2}{*}{ODEBlock1}     & Conv1  & 36,864   & (10.0\%)          & 4,672   & (1.3\%)            \\
                                           & Conv2  & 36,864   & (10.0\%)          & 4,672   & (1.3\%)            \\ \hline
            \multirow{3}{*}{Downsampling1} &   Conv & 8,192    & (100.0\%)         & 8,192   & (100.0\%)          \\
                                           & Conv1  & 73,728   & (100.0\%)         & 73,728  & (100.0\%)          \\
                                           & Conv2  & 147,456  & (100.0\%)         & 147,456 & (100.0\%)          \\ \hline
            \multirow{2}{*}{ODEBlock2}     & Conv1  & 147,456  & (10.0\%)          & 17,536  & (1.2\%)            \\
                                           & Conv2  & 147,456  & (10.0\%)          & 17,536  & (1.2\%)            \\ \hline
            \multirow{3}{*}{Downsampling2} &   Conv & 32,768   & (100.0\%)         & 32,768  & (100.0\%)          \\
                                           & Conv1  & 294,912  & (100.0\%)         & 33,920  & (11.5\%)           \\
                                           & Conv2  & 589,824  & (100.0\%)         & 67,840  & (11.5\%)           \\ \hline
            \multirow{2}{*}{ODEBlock3}     & Conv1  & 589,824  & (10.0\%)          & 67,840  & (1.2\%)            \\
                                           & Conv2  & 589,824  & (10.0\%)          & 67,840  & (1.2\%)            \\ \hline
            Others                         &        & 1,664    &                   & 1,664   &                    \\ \hline
            Total                          &        & 2,696,832& (16.2\%)          & 545,664 & (3.3\%)            \\ \hline
        \end{tabular}
    \end{minipage}
\end{table*}

\begin{figure}[ht]
    \centering
    \includegraphics[keepaspectratio, width=1\linewidth]{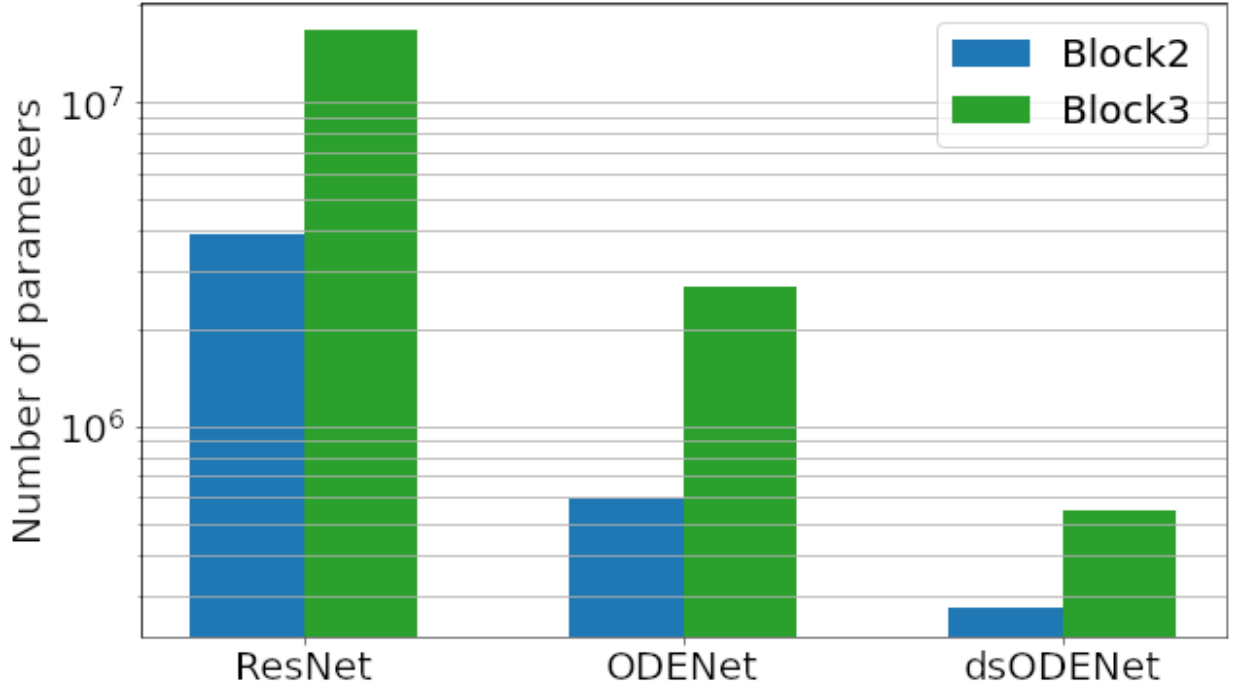}
    \caption{Comparison of the numbers of parameters of three models}
    \label{fig:parameters}
\end{figure}
A comparison of the number of parameters in the two building-block case (denoted as Block2) and the three building-block case (denoted as Block3) for ResNet, ODENet, and dsODENet is shown in Figure \ref{fig:parameters} in a logarithmic scale.
As shown, the number of parameters is greatly reduced in the order of ResNet, ODENet, and dsODENet.
Also, the three ODEBlocks case in dsODENet has fewer parameters than the two ODEBlocks case in ODENet.

\subsection{Accuracy}\label{subsec:accuracy}
For accuracy evaluations, 
Office-31 dataset (Office-31 \cite{office-31}),
road signs datasets (Synth signs \cite{synthsigns} and GTSRB \cite{gtsrb}),
and digit datasets (SVHN \cite{svhn} and MNIST \cite{mnist}) are
used as the datasets.
Their input image sizes are 256$\times$256, 40$\times$40, and 
32$\times$32, respectively.
The numbers of their classes are 31, 43, and 10.
The amounts of their domain shifts are various, small, and large.
The MNIST images are grayscale, while SVHN images are RGB colored; thus
the MNIST images are duplicated for three channels so that they are
compatible with the 3-channel SVHN images.
Office-31 is a popular dataset used for domain adaption tasks.
It contains 4,110 images, and they are divided into three domains:
Amazon (A-domain), Webcam (W-domain), and DSLR (D-domain).
The numbers of their images are 2817, 795, and 498, respectively.
A$\rightarrow$W means a domain adaptation scenario in which A-domain
is a source domain and W-domain is a target domain.
The numbers of W-domain and D-domain images are smaller than that of
A-domain.
A$\rightarrow$W and A$\rightarrow$D scenarios are examined in this paper because
domain adaptation from a domain with more images to that with less
images is a typical use case.
The edge domain adaptation procedure proposed in Section
\ref{sec:proposal} is used.
All the labeled source domain data and all the unlabeled target domain
data are used for the training phase.
The accuracy is then evaluated with all the labeled target data.

Either SGD or Adam (whichever shows better accuracy) is selected as an optimizer.
The learning rate is reduced based on Equation \ref{eq:learning_rate_decay}.
\begin{align}\label{eq:learning_rate_decay}
    \eta = \frac{\eta_0}{(1+\alpha p)^\beta} ,
\end{align}
where $\eta_0 = 0.01$, $\alpha = 10$, $\beta = 0.75$, and $p$ is
linearly changed from 0 to 1.
In the evaluations, the same experiments are executed three times and
their accuracy values are averaged and reported.
As for a teacher model, an initial model was pre-trained with ImageNet dataset, and using this initial model without the final fully-connected layer, the teacher model is trained.
The learning rate is reduced to 1/10 for these pre-trained layers.
Different student models that use ResNet-50, ODENet, and dsODENet are
compared in terms of accuracy.
ODENet and dsODENet use three ODEBlocks as shown in 
Figure \ref{fig:separable_ode3}.
The number of executions $C$ of an ODEBlock is set to 10.
They are also compared to a teacher model of ResNet-50 and other domain
adaptation techniques \cite{MobileDA,cdan,cdtrans,dadsn,DFA,adda,Mean-teacher}.

Table \ref{tbl:parameters_by_model} compares the parameter sizes of dsODENet, AlexNet, CDTrans, DSN, DFA, ADDA, and M-T.
Please note that the accuracies of CDTrans, DFA, and M-T are higher than the proposed approach as shown in Tables 5 - 7,
only the dsODENet and ADDA models could be implemented solely using BRAM/URAM slices of the target FPGA board (Xilinx ZCU104 board) due to the their scarcity.
Please note tha, CDTranst DSN, DFA, ADDA, and M-T are domain adaptation methods but they do not rely on the knowledge distillation;
thus they do not use student models.

\begin{table}[ht]
    \centering
    \caption{Number of parameters of models}
    \label{tbl:parameters_by_model}
    \begin{tabular}{l|r} \hline
        Mode                    & Number of parameters  \\ \hline
        dsODENet                &    595,072            \\
        AlexNet                 & 61,100,840            \\
        CDTrans \cite{cdtrans}  & 86,578,408            \\
        DSN \cite{dadsn}        &  1,102,159            \\
        DFA \cite{DFA}          & 31,493,150            \\
        ADDA \cite{adda}        &    426,070            \\
        M-T \cite{Mean-teacher} &  4,338,283            \\ \hline
    \end{tabular}
\end{table}

\subsubsection{Office-31 Dataset}\label{subsubsec:accuracy_office-31}
As a counterpart, MobileDA uses AlexNet as a student model and
ResNet-50 as a teacher model.
80\% of target domain data is used for the training.
As another counterpart, CDAN \cite{cdan} is also considered, which is a domain adaptation technique based on adversarial training.
In addition, CDTrans \cite{cdtrans} is a domain adaptation method using a transformer.
Table \ref{tbl:office-31_accuracy} shows the evaluation results of
CDAN, MobileDA, and our approach with different student models.

\begin{table}[ht]
    \centering
    \caption{Domain adaptation accuracy of Office-31 dataset}
    \label{tbl:office-31_accuracy}
    \begin{tabular}{l|cccccc} \hline
        Model                    & A$\rightarrow$W (\%)& A$\rightarrow$D (\%) \\ \hline
        CDAN \cite{cdan}         & 77.9  & 75.1  \\
        MobileDA \cite{MobileDA} & 71.5  & 75.3  \\
        CDTrans  \cite{cdtrans}  & 97.0  & 96.7  \\ \hline
        Teacher model ResNet-50  & 75.8  & 78.3  \\
        Student model1 ResNet-50 & 80.6  & 80.9  \\
        Student model1 ODENet    & 71.3  & 78.8  \\ \hline
        Student model1 dsODENet  & 80.4  & 79.1  \\
        Student model2 dsODENet  & 83.2  & 79.1  \\ \hline
    \end{tabular}
\end{table}
``Student model1 dsODENet'' and ``Student model2 dsODENet'' are our proposed models, where dsODENet is used for student model1 and student model2, respectively.
As shown in the table, the accuracy of A$\rightarrow$W is improved by 4.6\% and 2.8\% for the teacher model to ``Student model1 dsODENet'' and ``Student model1 dsODENet'' to ``Student model2 dsODENet'', respectively.
Also, the accuracy of A$\rightarrow$D is improved by 0.8\% for the teacher model to ``Student model1 dsODENet''.
However, there is no improvement in accuracy from the student model1
to the student model2.
Please note that the proposed approach differs from the original MobileDA in the following points.
The proposed approach does not use hard target loss in the loss function because in our evaluation environment the accuracy was slightly improved by not using it.
In the proposed approach, two student models are used.
That is, a student model1 is trained from a teacher model, and then a student model2 is trained from the student model1.
In this paper, we assume that ``Student model1 dsODENet'' corresponds to the results of MobileDA when dsODENet is used as a student model.
In other words, accuracy differences between ``Student model1 dsODENet'' and ``Student model2 dsODENet'' are improvements obtained by the proposed approach.
As shown in Tables 5, 6, and 7, the proposed approach actually improves the accuracy in most cases.
Accuracy of CDTrans outperforms the proposed approach  while its parameter size is much larger than the proposed one.

\begin{figure}[ht]
    \centering
    \includegraphics[keepaspectratio, width=1\linewidth]{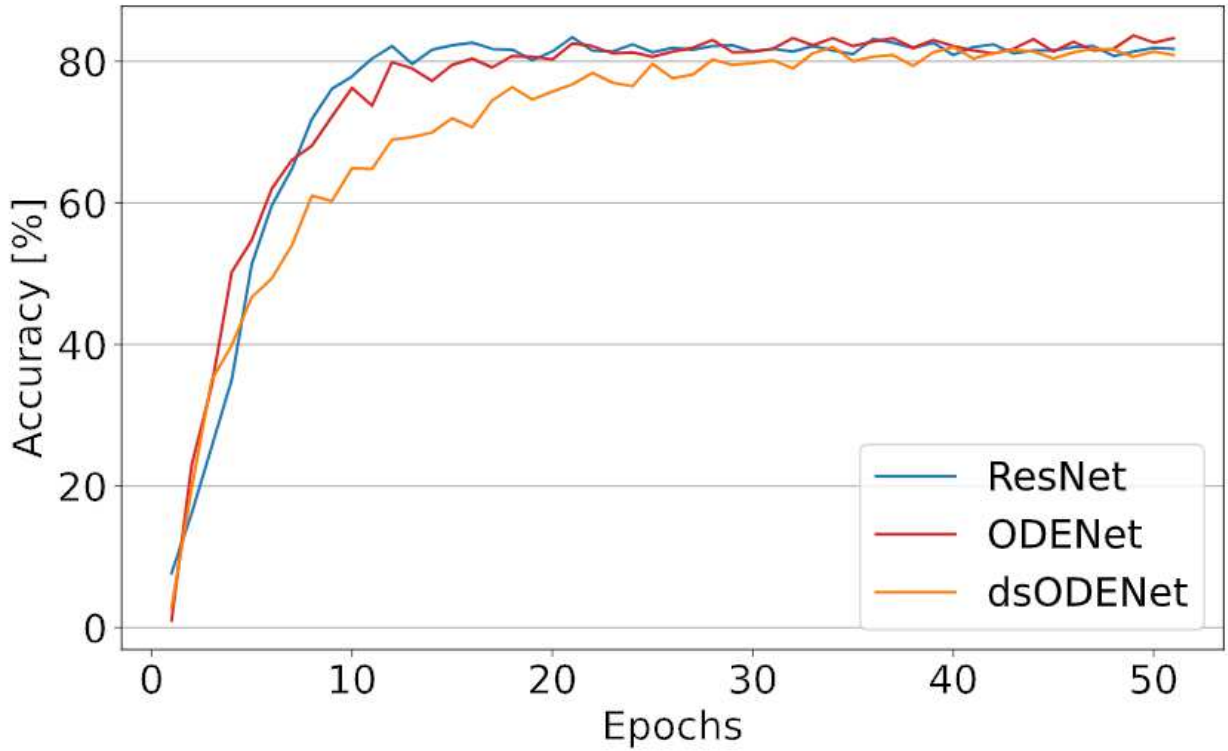}
    \caption{Training speed of different student models in Offce-31 dataset}
    \label{fig:office_a_d}
\end{figure}
Figure \ref{fig:office_a_d} shows the training speeds of different
student models (ResNet-50, ODENet, and dsODENet) for A$\rightarrow$D
scenario.
As shown in the figure, the student model of ResNet-50 is converged
faster than the others, followed by those of dsODENet and ODENet.
In ResNet-50, the number of implemented building blocks is optimized
for each feature map size in this experiment.
These numbers in ResNet-50 are interpreted as the numbers of continuous
executions of ODEBlocks in the cases of ODENet and dsODENet
($C$ in Figures \ref{fig:separable_ode2} and \ref{fig:separable_ode3}).
As mentioned, in ODENet and dsODENet, a single execution of an
ODEBlock is interpreted as a single step of Euler method.
Reducing the number of executions $C$ of an ODEBlock degrades the
approximation performance and reduces the accuracy.
To obtain a stable accuracy, the number of continuous
executions $C$ of an ODEBlock is equally set to 10 for each feature
map size in ODENet and dsODENet cases, though
there is still a room for optimization.
Please note that the training speed of dsODENet is faster than ODENet,
because in dsODENet the number of parameters to be trained is reduced
by 54.2\% to 79.8\% by DSC.

\subsubsection{Road Sign Dataset}\label{subsubsec:accuracy_traffic_signs}
As a counterpart, DSN \cite{dadsn} which is a domain adaptation based
on adversarial training is compared to our approach with different
student models.
In addition, DFA \cite{DFA} employs a latent alignment approach based on the encoder-decoder formulation.
Table \ref{tbl:mark_accuracy} shows the evaluation results of DSN and
our approach with different student models.
Please note that network structures of DSN and ours are different and
a fair comparison is difficult.
Accuracy of DFA is higher than that of the proposed approach and DSN while its parameter size is larger than the proposed one.
\begin{table}[ht]
    \centering
    \caption{Domain adaptation accuracy of road sign datasets}
    \label{tbl:mark_accuracy}
    \begin{tabular}{l|c} \hline
        Model                   & Synth signs$\rightarrow$GTSRB (\%) \\ \hline
        DSN \cite{dadsn}        & 93.1\\
        DFA \cite{DFA}          & 97.5\\ \hline
        Teacher model ResNet-50 & 95.1\\
        Student model1 ResNet-50 & 97.1\\
        Student model1 ODENet    & 97.0\\ \hline
        Student model1 dsODENet  & 97.1\\
        Student model2 dsODENet  & 97.3\\ \hline
    \end{tabular}
\end{table}

``Student model1 dsODENet'' and ``Student model2 dsODENet'' are our proposed models, where dsODENet is used for student model1 and student model2, respectively.
As shown in the table, the accuracy of Synth signs$\rightarrow$GTSRB is improved by 2.0\% and 0.2\% for the teacher model to ``Student model1 dsODENet'' and ``Student model1 dsODENet'' to ``Student model2 dsODENet'', respectively.
The accuracy of ``Student model1 dsODENet'' becomes higher than that of
the teacher model because the two datasets contain a large enough
number of images and their domain shift is small.
In contrast, the improvement in accuracy from 
``Student model1 dsODENet'' to ``Student model2 dsODENet'' is small.
As the domain shift is small, the accuracies of ResNet-50, ODENet, and
dsODENet are similar.

\subsubsection{Digit Dataset}\label{subsubsec:accuracy_digit}
As a counterpart, ADDA \cite{adda} which is also a domain adaptation
based on adversarial training is considered.
In addition, Mean teacher (M-T) \cite{Mean-teacher} is an approach using a self-ensembling technique.
    
Table \ref{tbl:digits_accuracy} shows the evaluation results of ADDA
and our approach with different student models.
\begin{table}[ht]
    \centering
    \caption{Domain adaptation accuracy of digit datasets}
    \label{tbl:digits_accuracy}
    \begin{tabular}{l|c} \hline
        Model                    & SVHN$\rightarrow$MNIST (\%) \\ \hline
        ADDA \cite{adda}         & 76.0\\ \hline
        M-T \cite{Mean-teacher}  & 93.3\\ \hline 
        Teacher model ResNet-50  & 76.5\\
        Student model1 ResNet-50 & 82.6\\
        Student model1 ODENet    & 82.5\\ \hline
        Student model1 dsODENet  & 83.5\\
        Student model2 dsODENet  & 86.6 \\ \hline
    \end{tabular}
\end{table}
As shown in the table, there is an improvement of 7.0\% and 1.7\% in accuracy from teacher model to ``Student model1 dsODENet'' and from ``Student model1 dsODENet'' to ``Student model2 dsODENet''.
Accuracy of M-T is higher than that of the proposed approach and ADDA while its parameter size is larger than the proposed one.
We consider that this two-step improvement in accuracy is the result of both the domain adaptation and knowledge distillation.
As shown in Algorithm \ref{alg:prop_method}, target domain samples to be used for knowledge distillation are selected by a given threshold value.
If the threshold value is low, the number of training samples for a student model1/2 increases,
while it becomes more likely that samples mislabeled by a teacher model/student model1 are used during training.
We set a higher threshold value in the first knowledge distillation from a teacher model to a student model1.
We observed a lower threshold causes a training instability because this is an adaptation from a larger model to a smaller model.
We set a lower threshold value in the second knowledge distillation from the student model1 to a student model2.
Since their model sizes are the same, we observed that the lower threshold can increase the number of training samples and produce slightly better accuracy, as reported in this section.
Please note that adding the second-step is meaningful while adding the third-step does not have a similar impact if the threshold value is the same.

\begin{figure}[ht]
    \centering
    \includegraphics[keepaspectratio, width=1\linewidth]{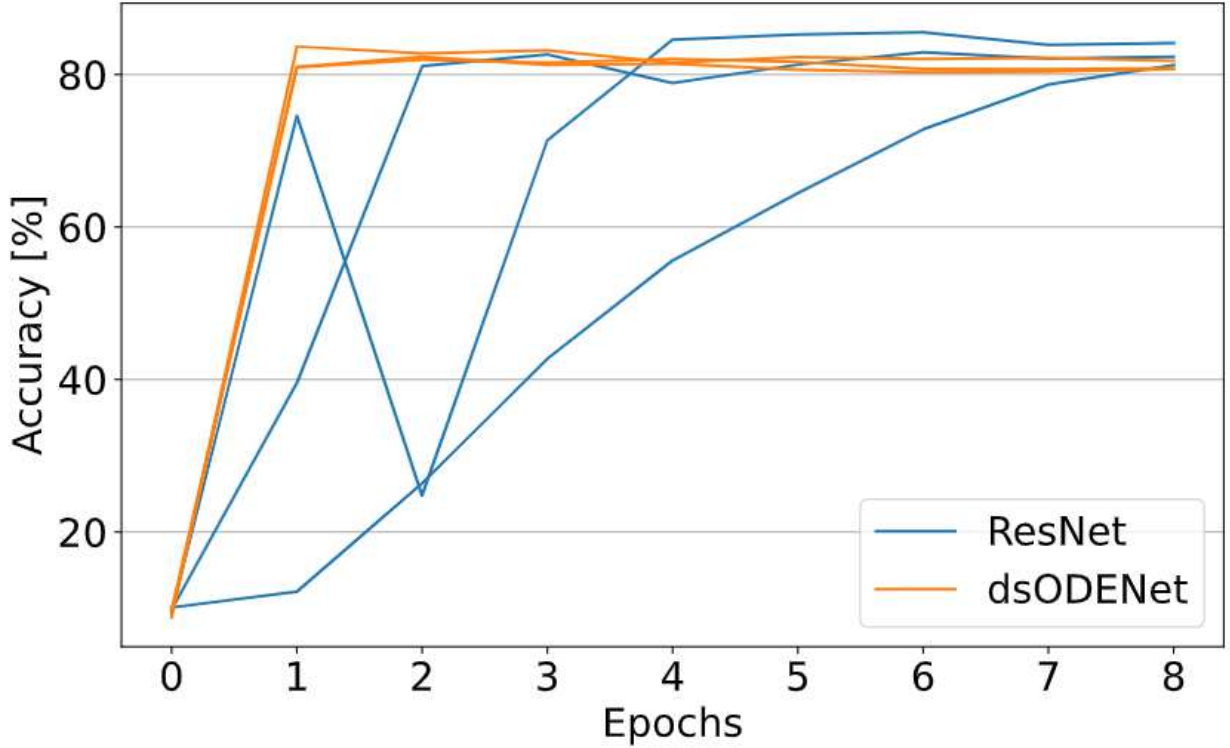}
    \caption{Training speed of different student models in digit datasets}
    \label{fig:digits_r_s}
\end{figure}

Figure \ref{fig:digits_r_s} shows the training speeds of different
student models (ResNet-50 and dsODENet) for SVHN$\rightarrow$MNIST
scenario.
In Figure \ref{fig:digits_r_s}, six lines are shown in total since
ResNet-50 and dsODENet are trained three times each.
dsODENet is converged fast and stable compared to ResNet-50 because the
number of parameters of dsODENet is significantly reduced compared to
ResNet-50.
For domain adaptation tasks with a larger domain shift, the accuracy of
ResNet-50 becomes higher than that of dsODENet but the training curve
of dsODENet is still stable.

\subsection{FPGA Evalutation}\label{subsec:area_speed}
%
We develop a dsODENet accelerator by offloading the computation onto the FPGA SoC. 
Since a quantization is applied in our FPGA implementation, in this section it is evaluated in terms of the inference accuracy, resource utilization, and performance on the FPGA.

\subsubsection{Effect of Quantization}\label{subsubsec:quantization}
%
In the FPGA implementation, a quantization is applied to pack the weight parameters and feature maps in the BRAM and URAM slices.
More specifically, a 20-bit fixed-point format is used in the convolutional layers, while a 24-bit fixed-point format is used in the other layers.
It is thus expected that the actual accuracy of the FPGA implementation slightly differs from those evaluated in Section \ref{subsec:accuracy}. 
Table \ref{tbl:quantization_digits_accuracy} shows evaluation results of the accuracy with digit Dataset when the quantization is applied.
We evaluated the effect of quantization only with the digit datasets.
This is because only the dsODENet models for relatively small inputs (digit images) could be implemented solely using BRAM/URAM slices due to their scarcity.
Because the other datasets (e.g., Office-31) contain larger images,
which necessitates more aggressive quantizations and manual optimizations in order to implement dsODENet models for those larger inputs on the target FPGA.
Since BRAM/URAM sizes of FPGA devices are continually increasing, such optimized implementations for larger datasets are our future work.
The results show that the accuracy drops by 0.02\% but this accuracy loss is much smaller than the accuracy improvement obtained by the proposed domain adaptation method compared to the original teacher model.
\setlength{\tabcolsep}{1mm}
\begin{table}[ht]
    \centering
    \caption{Domain adaptation accuracy of digit datasets with quantization}
    \label{tbl:quantization_digits_accuracy}
    \begin{tabular}{l|c} \hline
        Model                    & {\scriptsize SVHN$\rightarrow$MNIST (\%)}\\ \hline
        {\scriptsize Teacher model ResNet-50}  & 76.51\\  
        {\scriptsize Student model2 dsODENet w/o quantization} & 86.57 \\
        {\scriptsize Student model2 dsODENet w/ quantization}  & 86.55 \\ \hline
    \end{tabular}
\end{table}

\subsubsection{Resource Utilization}\label{subsubsec:resource_utilization}
Table 8 shows FPGA resource utilizations when two ODEBlocks case (Block2) and three ODEBlocks case (Block3) are implemented.
Importantly, weight parameter and feature map arrays of the Block3 case can be implemented on the BRAM and URAM slices of the FPGA without using external DRAMs. 
The results also show that the Block2 implementation still has room in the BRAM and URAM capacities, while they are close to 100\% in the Block3 implementation, 
which means that there is room for improving the performance and accuracy in the Block2 implementation by introducing more aggressive parallelization and wider bit widths.

\begin{table}[ht]
    \centering
    \caption{FPGA resource utilization of dsODENet}
    \label{tbl:area1_source}
    \begin{tabular}{l|ccccc} \hline
                 & BRAM & DSP & FF & LUT & URAM\\ \hline
        Block2   & 208 (67\%)& 760 (44\%)    & 22,880 (5\%) & 63,461 (28\%) & 64 (67\%)\\
        Block3   & 265 (85\%)& 138 (8\%)     & 19,755 (4\%) & 49,014 (21\%) & 92 (96\%)\\ \hline
        \end{tabular}
\end{table}


\subsubsection{Execution Time}\label{subsubsec:excecution_time}
Table \ref{tbl:layer_time} compares the execution time of each block for a single data sample between our FPGA implementation and its software counterpart (denoted as FPGA and CPU). 
The mean and standard deviation of the execution time to infer one image is calculated by inferring 100 images 30 times.
The execution time for each block in the FPGA is calculated from the number of cycles of the high-level synthesis result.
In the FPGA case, a DMA transfer between PS–PL is additionally required, while the high-level synthesis result takes into account only for the PL part.
Thus, there is an execution time difference between the high-level synthesis result and the actual on-board execution time that takes into account the DMA transfer in the PS part, which is denoted as DMA transfer (PS) in Table \ref{tbl:layer_time}.
Both the ODEBlocks and downsampling blocks were accelerated by 3.1-82.0 times, which contributes to the overall speedup of 23.8 times and 30.6 times with and without considering the DMA transfer (PS).
If we focus on a single execution time of each block, the downsampling blocks achieved better speedups than ODEBlocks, as they involve normal convolution, which is more compute-intensive than DSC.
However, ODEBlocks are repeatedly executed $C$ times and thus their impact is higher than those of Downsampling1 and Downsampling2, as shown in Table \ref{tbl:layer_time}.
Further performance improvement of the FPGA can be expected by more aggressive parallelization of multiply-add operations.

Please note that an advantage of our approach is that dsODENet is a ResNet-like backbone architecture which can be stored in small but high-throughput on-chip memories of FPGAs.
It is thus orthogonal to other promising techniques, such as split-CNN technique \cite{FMSCNN} and aggressive quantization including BNNs \cite{FracBNN}.
Another advantage of our approach is that, by increasing $C$ parameter of ODENet/dsODENet, we can increase the number of iterations of ODEBlocks/dsODEBlocks to improve the accuracy without increasing the number of parameters.
\begin{table}[ht]
    \centering
    \caption{Execution time of each block on FPGA}
    \label{tbl:layer_time}
    \begin{tabular}{l|r|r|r} \hline
                          & FPGA (ms) & CPU (ms) & Speedup\\ \hline
        ODEBlock1         & 7.30      & 112.85   & 15.5 \\
        Downsampling1     & 5.14      & 16.13    & 3.1 \\
        ODEBlock2         & 6.42      & 213.42   & 33.2 \\
        Downsampling2     & 1.46      & 34.56    & 23.7 \\
        ODEBlock3         & 4.84      & 396.98   & 82.0 \\
        DMA transfer (PL) & 0.13      & -        & -    \\
        DMA transfer (PS) & 7.24      & -        & -    \\ \hline
        Total             & 32.53 $\pm$ 1.58 & 773.94 $\pm$ 35.0  & 23.8 \\ \hline
        \end{tabular}
\end{table}


\section{Summary}\label{sec:conc}
In this paper, a combination of Neural ODE and DSC, called dsODENet,
is proposed and implemented for FPGAs.
dsODENet is applied to a distillation-based edge domain adaptation as
student models.
All the dsODENet blocks except the pre- and post-processing layers are
implemented on PL part of Xilinx ZCU104 FPGA board and the others
are executed on PS part.
Importantly, all their parameter and feature map arrays are stored in
URAM and BRAM instances of the FPGA without relying on external DRAMs.
It is evaluated in terms of domain adaptation accuracy, training
speed, FPGA resource utilization, and speedup rate compared to a
software execution.
Regarding the domain adaptation accuracy, the student models of dsODENet
are comparable to or better than that of ODENet and better than
some existing domain adaptation methods.
The total parameter size of dsODENets without pre- and post-processing
layers is reduced by 54.2\% to 79.8\%.
The FPGA implementation accelerates the prediction tasks by 23.8
times than a software implementation running on PS part.


As a future work, we are optimizing our FPGA implementations
by introducing 16-bit fixed-point or bfloat16 representations to
suppress the on-chip memory utilizations and enable more aggressive
parallel implementation.


\bibliographystyle{unsrt}
\bibliography{refer}

\end{document}